\documentclass[sigconf]{acmart}
\AtBeginDocument{%
  }

\setcopyright{cc}
\setcctype{by}
\copyrightyear{2025}
\acmYear{2025}

\acmConference[MMAsia '25 Workshops]{ACM Multimedia Asia Workshops}{December 9--12, 2025}{Kuala Lumpur, Malaysia}
\acmBooktitle{ACM Multimedia Asia Workshops (MMAsia '25 Workshops), December 9--12, 2025, Kuala Lumpur, Malaysia}
\acmDOI{10.1145/3769748.3773355}
\acmISBN{979-8-4007-2247-9/2025/12}

\usepackage{subcaption}
\begin{document}

\title{Estimation of Fireproof Structure Class and Construction Year for Disaster Risk Assessment}

\author{Hibiki Ayabe}
\email{hibiki-ayabe@uec.ac.jp}
\orcid{0009-0009-2801-6892}
\affiliation{%
  \institution{The University of Electro-Communications}
  \city{Chofu}
  \state{Tokyo}
  \country{Japan}
}

\author{Kazushi Okamoto}
\email{kazushi@uec.ac.jp}
\orcid{0000-0002-9571-8909}
\affiliation{%
  \institution{The University of Electro-Communications}
  \city{Chofu}
  \state{Tokyo}
  \country{Japan}
}

\author{Koki Karube}
\email{karubekoki@uec.ac.jp}
\orcid{0009-0005-2699-1193}
\affiliation{%
 \institution{The University of Electro-Communications}
 \city{Chofu}
 \state{Tokyo}
 \country{Japan}
}

\author{Atsushi Shibata}
\email{shibata-atsushi@uec.ac.jp}
\orcid{0000-0003-1794-8562}
\affiliation{%
  \institution{The University of Electro-Communications}
  \city{Chofu}
  \state{Tokyo}
  \country{Japan}
}

\author{Kei Harada}
\email{harada@uec.ac.jp}
\orcid{0009-0009-5453-6699}
\affiliation{%
  \institution{The University of Electro-Communications}
  \city{Chofu}
  \state{Tokyo}
  \country{Japan}
}


\begin{abstract}
Structural fireproof classification is vital for disaster risk assessment and insurance pricing in Japan.
However, key building metadata such as construction year and structure type are often missing or outdated, particularly in the second-hand housing market.
This study proposes a multi-task learning model that predicts these attributes from facade images.
The model jointly estimates the construction year, building structure, and property type, from which the structural fireproof class--defined as H (non-fireproof), T (semi-fireproof), or M (fireproof)--is derived via a rule-based mapping based on official insurance criteria.
We trained and evaluated the model using a large-scale dataset of Japanese residential images, applying rigorous filtering and deduplication.
The model achieved high accuracy in construction-year regression and robust classification across imbalanced categories.
Qualitative analyses show that it captures visual cues related to building age and materials.
Our approach demonstrates the feasibility of scalable, interpretable, image-based risk-profiling systems, offering potential applications in insurance, urban planning, and disaster preparedness.
\end{abstract}




\maketitle

\section{Introduction}\label{sec:introduction}

Natural disasters continue to pose significant risks to residential properties in Japan~\cite{oecd2021catastorophe}.
Fire and earthquake insurance systems have been developed to mitigate financial losses, and the attendant insurance premiums are determined based on various building attributes such as the type of structure and construction year~\cite{sheehan2023benefits}.
In Japan, a unique structural fireproof classification system--categorized as non-fireproof (H), semi-fireproof (T), and fireproof (M)--plays a crucial role in calculating fire and earthquake insurance premiums.
The classification is officially defined and adopted by Japanese insurers in accordance with guidelines issued by the General Insurance Rating Organization of Japan~\cite{GIROJ2024}.
However, such building metadata, including construction year are often unavailable or outdated, especially in the second-hand housing markets, because the records are less systematically maintained or updated by either owners or public authorities.
This lack of accessible and reliable data--especially regarding attributes critical for insurance assessment--makes it difficult for stakeholders such as homeowners, buyers, and insurers to accurately evaluate disaster risks.
As a result, housing risk information remains fragmented and opaque, limiting the ability of individuals to make informed insurance decisions and insurers to appropriately price risks.

This study aims support informed decision-making by individuals, governments, and insurers through increased transparency in disaster-related building risks.
Key building attributes can be estimated from exterior images that are relevant to disaster vulnerability.
This study focuses on predicting the structural fireproof category and construction year, which are central to insurance risk assessment in Japan.
Recent advances in computer vision and deep learning enabled automated analysis of building attributes from facade images~\cite{benz2023automated,chen2022attribute,dionelis2025age,li2018age,you2017appraisal,ogawa2023buildings,liang2025openfacades,blier-wong2024representation,murdoch2024residential, ayabe2025age}.
This study proposes a multi-task learning (MTL) framework for predicting building vulnerability attributes from exterior images using an attribute-based approach.
A model that simultaneously predicts the construction year, building structure, and property type is developed.
It applies a rule-based mapping to derive the structural fireproof classification (H, T, M), which is a key determinant in insurance risk assessment in Japan, by using the predicted building structure and property type.
This attribute-based formulation offers interpretability and practical alignment with insurance standards, as the intermediate predictions (e.g., ``concrete'' vs. ``wooden'' structure, or ``communal'' vs. ``detached'') reflect tangible characteristics used in formal assessments.

This study addresses the following research questions:
\begin{itemize}
  \item[\textbf{RQ1}.]{Can disaster-relevant building attributes--including construction year, building structure, and property type--be accurately estimated from residential facade images using multi-task learning?}
  \item[\textbf{RQ2}.]{How effectively can the fireproof class be inferred from intermediate attribute predictions, such as building structure and property type, in a hierarchical modeling approach?}
\end{itemize}
To answer RQ1, we trained a multi-task learning model on labeled facade images to jointly predict the construction year, structure type, and property type based on visual features.
To answer RQ2, we conducted ablation experiments to evaluate how intermediate predictions affect fireproof classification accuracy and assess the scalability of the hierarchical approach in practical risk profiling.

The novelty of this study lies in the design of a multi-task learning framework tailored for disaster-relevant building attributes, combined with preprocessing strategies for a large-scale, noisy facade image dataset.
\section{Literature Review}\label{sec:related_work}

\subsection{Real Estate Property Analysis Based on Deep Learning}
Traditional real estate appraisal and classification depend heavily on manual data collection and metadata are provided by property owners or listing platforms. 
However, computer vision approaches that predict property attributes directly from images have emerged in recent years ~\cite{benz2023automated,chen2022attribute,dionelis2025age,li2018age,you2017appraisal,ogawa2023buildings,liang2025openfacades,blier-wong2024representation,murdoch2024residential, ayabe2025age}.
You et al.~\cite{you2017appraisal} estimated rental amounts of residential properties from facade images.
Li et al.~\cite{li2018age}, Benz et al.~\cite{benz2023automated}, and Dionelis et al.~\cite{dionelis2025age} proposed methods to infer construction years from facade images. 
Ogawa et al.~\cite{ogawa2023buildings} predicted construction years and building structures from building images.

This was further extended to disaster risk assessment by Chen et al.~\cite{chen2022attribute}; they estimated flood vulnerability by leveraging structural attributes inferred from exterior images. 
Similarly, Hafidz et al.~\cite{hafidz2024typology} analyzed seismic vulnerability based on facade-derived typological information.
Blier-Wong et al.~\cite{blier-wong2024representation} developed image-based representations for insurance-premium estimation.
Japan faces particularly high disaster risks, especially from earthquakes and subsequent fires, owing to its geographical and urban characteristics~\cite{oecd2021catastorophe}.
Despite these advances elsewhere, such applications remain underexplored in Japan. 
In particular, studies related to fireproof classification or disaster risk estimation by analyzing the exterior image are scarce. 
Therefore, this study uses an MTL model to estimate key disaster-relevant attributes--construction year, building structure, property type, and fireproof classification--from exterior images, while noting that direct empirical comparisons with prior works are difficult due to biased sampling in existing datasets and differences in data distributions.

\subsection{Multi-task Learning in Vision Tasks}
MTL, which aims to train models that simultaneously perform multiple tasks, has been shown to improve generalization and data efficiency, particularly in computer vision~\cite{caruana1997multitask, zamir2018taskonomy}. 
For instance, Zamir et al.~\cite{zamir2018taskonomy} demonstrated that related visual tasks--such as semantic segmentation and depth estimation--can benefit from shared representations through task transfer analysis.
However, MTL remains underexplored in real estate and disaster risk assessment. 
One of the few studies, Chen et al.~\cite{chen2022attribute}, applied MTL to predict structural attributes such as height, type, and floor area, using building images. 
Despite growing interest in vision-based risk assessment, the use of MTL to jointly predict disaster-relevant attributes from facade images--especially fireproof classification--is limited in existing literature.

To handle the challenge of balancing heterogeneous tasks, Kendall et al.~\cite{kendall2018multitask} proposed an uncertainty-based loss-weighting strategy, which enables adaptive learning across regression and classification objectives. 
Building on this foundation, we propose an MTL framework that simultaneously predicts construction year, building structure, and property type.
The predicted building structure and property type are then used to infer the fireproof classification through a rule-based mapping. 
This mapping, grounded in intermediate attribute estimation, improves interpretability, aligns with real-world insurance definitions, and enables scalable disaster-vulnerability assessment from facade imagery.
\section{Experiment}\label{sec:method}

\subsection{Tasks and Evaluation Metrics}
This study evaluates a multi-task learning model trained to predict multiple building attributes relevant to disaster risk assessment.
The following four tasks are defined:
\begin{itemize}
\item[\textbf{Task 1}]{Construction year prediction (regression)}
\item[\textbf{Task 2}]{Building structure prediction (classification)}
\item[\textbf{Task 3}]{Property type prediction (classification)}
\item[\textbf{Task 4}]{Fireproof structural class prediction (classification).}
\end{itemize}
To evaluate model performance across these heterogeneous tasks, this study adopted task-specific metrics that capture both predictive accuracy and robustness to class imbalance.

For Task 1 (construction year prediction), performance was assessed using mean absolute error (MAE), root mean squared error (RMSE), and median absolute error (MedAE). 
While MAE and RMSE reflect overall prediction accuracy, MedAE is less sensitive to outliers—an important consideration given the temporal skew of the dataset toward more recent buildings.
For the classification tasks--Tasks 2, 3, and 4--evaluation metrics included accuracy and macro-averaged F1-score, as well as per-class F1 to address class imbalance. 
Confusion matrices were also analyzed to identify patterns of systematic misclassification, particularly in Task 4, where the final fireproof class was inferred through a hierarchical process based on intermediate attributes.

\subsection{Dataset}
\begin{figure*}[t]
  \centering
  \begin{subfigure}[b]{0.18\linewidth}
    \includegraphics[width=\linewidth]{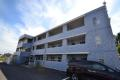}
    \caption{Communal, Concrete}
    \label{fig:communal_concrete}
  \end{subfigure}
  \hfill
  \begin{subfigure}[b]{0.18\linewidth}
    \includegraphics[width=\linewidth]{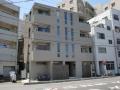}
    \caption{Communal, Steel}
    \label{fig:communal_steel}
  \end{subfigure}
  \hfill
  \begin{subfigure}[b]{0.18\linewidth}
    \includegraphics[width=\linewidth]{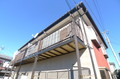}
    \caption{Communal, Wooden}
    \label{fig:communal_wooden}
  \end{subfigure}
  \hfill
  \begin{subfigure}[b]{0.18\linewidth}
    \includegraphics[width=\linewidth]{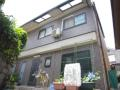}
    \caption{No-communal, Steel}
    \label{fig:nocomm_steel}
  \end{subfigure}
  \hfill
  \begin{subfigure}[b]{0.20\linewidth}
    \includegraphics[width=\linewidth]{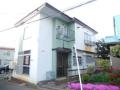}
    \caption{No-communal, Wooden}
    \label{fig:nocomm_wooden}
  \end{subfigure}
    \caption{Representative facade samples from five combinations of \textit{property type} and \textit{building structure}. No sample was available for the combination of \textit{No-communal} and \textit{Concrete-like}.}
  \label{fig:qualitative_by_group}
\end{figure*}

This study used the LIFULL HOME'S dataset~\cite{LIFULL}, which includes metadata for approximately 5.33 million real estate properties in Japan, as well as approximately 83 million associated images such as floor plans and interior/exterior photos. 
We used approximately 9.5 million images, labeled ``exterior'', as the visual input to this model. 
Each image is linked to property-level metadata, including construction year, building structure, and property type, which are used as prediction targets.
Figure~\ref{fig:qualitative_by_group} shows examples of facade images.

\subsubsection{Preprocessing}
This study first extracted the metadata for each property by removing entries with missing values in the construction year, building structure, or property type.
Additionally, properties constructed before 1915 were excluded because the data were few and often lacked reliable records.
Next, to focus on residential buildings, we retained only data labeled as Apartment, House, Single-family house, Terrace house, Townhouse, Sublease, or Dormitory; we excluded commercial and non-residential properties from the dataset.

The dataset was then split into training and test sets at a ratio of 8:2 the property level to prevent data leakage.
After the split, 3.14 million unique training images were obtained after removing corrupted or unreadable image files.
In removing corrupted or unreadable image files, perceptual hashing (pHash)~\cite{lv2012phash} was applied to eliminate near-duplicate images.
In addition, CLIP~\cite{radford2021visual,feng2024clipcleaner}--classifying images into four categories (Entire residential property, No residential property, Inside of residential property, and Others)--was employed to retain only Entire residential property images.
Finally, this study used 2.77 million training images and 0.40 million test images.

\begin{figure}[t]
    \centering
    \includegraphics[width=0.85\linewidth]{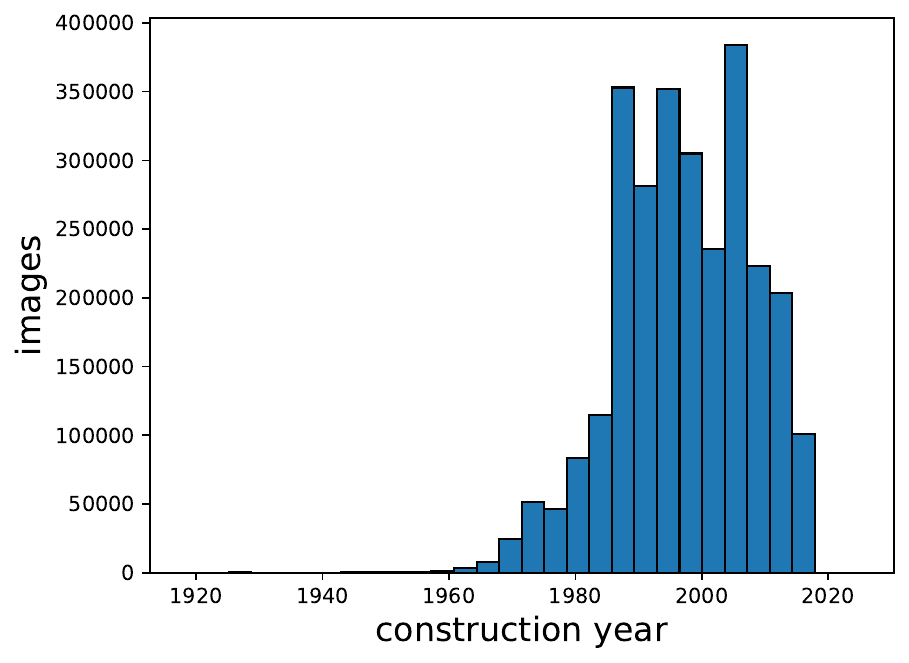}
    \caption{Distribution of construction years in the filtered dataset. The data is skewed toward more recent buildings.}
    \label{fig:construction_year_distribution}
\end{figure}

The distribution of construction years in the final dataset is shown in Figure~\ref{fig:construction_year_distribution}.
Notably, 43.9\% of the samples were built after 2000, while only 6.54\% were constructed before 1980, indicating a strong temporal imbalance.
This imbalance motivated the use of robust evaluation metrics and stratified error analysis in the construction-year estimation task.

\subsubsection{Label Construction}\label{sec:label-construction}
To define ground-truth labels for the structural fireproof classification task, we constructed a rule-based mapping derived from two metadata fields: building structure and property type~\cite{GIROJ2024}.
We defined three fireproof classes --H (non-fireproof), T (semi-fireproof), and M (fireproof)--based on rule-based mapping between building structure and property type, as summarized in Table~\ref{tab:fireproof_rules}.
In practice, fireproof classification for insurance purposes is determined through formal inspection and detailed architectural documentation. 
While our rule-based mapping may not perfectly reflect official classifications, it offers a practical and interpretable proxy grounded in real-world criteria.
This enables large-scale label generation and supports supervised learning aligned with insurance standards.

\begin{table}[t]
    \centering
    \caption{Rule-based mapping for fireproof classification.}
    \label{tab:fireproof_rules}
    \begin{tabular}{lll}
        \toprule
        Building Structure & Property Type & Fireproof Class \\
        \midrule
        Concrete-like & Any & M (Fireproof) \\
        Steel-like & Communal & M (Fireproof) \\
        Steel-like & Non-communal & T (Semi-fireproof) \\
        Wooden-like & Any & H (Non-fireproof) \\
        \bottomrule
    \end{tabular}
\end{table}

\subsection{Multi-task Model Design}
We implemented an MTL model based on ResNet-101~\cite{he2016residual}, which was pretrained on ImageNet~\cite{deng2009imagenet}.
The model shares a common convolutional backbone and branches into three output heads for the following tasks: (1) construction year (regression), (2) building structure (concrete-like, steel-like, and wooden-like), and (3) property type (communal and non-communal).
To balance the heterogeneous loss functions of the regression and classification tasks, the uncertainty-weighted loss formulation proposed by Kendall et al.~\cite{kendall2018multitask} was employed, which incorporates task-specific homoscedastic uncertainty:
\begin{equation}
\mathcal{L} = \sum_{i} \frac{1}{2 \sigma^{2}_{i}} \mathcal{L}_{i} + \log \sigma_{i}
\end{equation}
Where, $\mathcal{L}_{i}$ is the loss for each task (mean squared error for regression, cross-entropy for classification), and $\sigma_{i}$ is a learnable parameter representing task uncertainty.
These uncertainty parameters are optimized simultaneously with the model weights during training, allowing dynamic adjustment of task contributions based on their relative difficulty.
The model was trained using the Adam optimizer~\cite{kingma2017adam}. 
To evaluate the effect of learning rate on model convergence and generalization, two settings were employed in this experiment: $10^{-5}$ and $10^{-6}$.

\section{Result and Discussion}\label{sec:result_and_discussion}

\subsection{Multi-task Learning (Task 1, Task2, and Task3)}
\subsubsection{Construction Year Prediction (Task 1)}
Table~\ref{tab:year_estimation_results} presents the results for construction year prediction using the proposed MTL model. 
According to the table, the better-performing configuration was trained with a learning rate of $10^{-5}$.
The MAE, RMSE, and MedAE values indicate strong predictive performance across both central tendency and robustness to outliers.
The model that was trained with a lower learning rate of $10^{-6}$ exhibits moderately degraded performance.
There is a tendency that the higher learning rate facilitates better convergence in this multi-task setting.

While direct comparison is complex owing to differences in datasets and task definitions, these results are comparable to or exceed those reported in prior work on construction year estimation from facade images~\cite{li2018age}. 
Notably, this model achieved this performance in a multi-task context--jointly learning from related architectural attributes--without requiring additional handcrafted features or external data sources.
These findings support the effectiveness of MTL in leveraging shared visual representations, while also demonstrating the feasibility of year estimation from large-scale, in-the-wild image datasets such as LIFULL HOME'S dataset.

\begin{table}[t]
    \centering
    \caption{Performance comparison on construction year estimation. Lower values indicate better predictive accuracy.}
    \label{tab:year_estimation_results}
    \begin{tabular}{cccc}
        \toprule
        LR & MAE $\downarrow$ & RMSE $\downarrow$ & MedAE $\downarrow$ \\
        \midrule
        $10^{-5}$ & 4.97 & 6.88 & 3.59 \\
        $10^{-6}$ & 6.02 & 7.78 & 4.90 \\
        \bottomrule
    \end{tabular}
\end{table}

\subsubsection{Building Structure and Property Type Prediction (Task 2 and 3)}
The fireproof structural class is derived from two intermediate attributes: the building structure and the property type; therefore, accurate prediction of these components is crucial, as any misclassification may propagate to the fireproof structural category prediction (Task 4).
The classification performance of the proposed MTL model on these intermediate tasks was evaluated, and the results are summarized in Table~\ref{tab:intermediate_attribute_results}.
The model trained with a learning rate of $10^{-5}$ achieved a building structure classification accuracy of 92.75\% and a macro F1 score of 0.8714, indicating strong overall performance as well as robustness across imbalanced classes.
For property type, the same model obtained 83.21\% accuracy and a macro F1 of 0.7945.
A slight drop in performance was observed for the model trained with a learning rate of $10^{-6}$ , which suggests that the lower learning rate may have led to underfitting.

\begin{figure*}[t]
    \centering
    \begin{subfigure}[b]{0.27\textwidth}
        \centering
        \includegraphics[width=\linewidth]{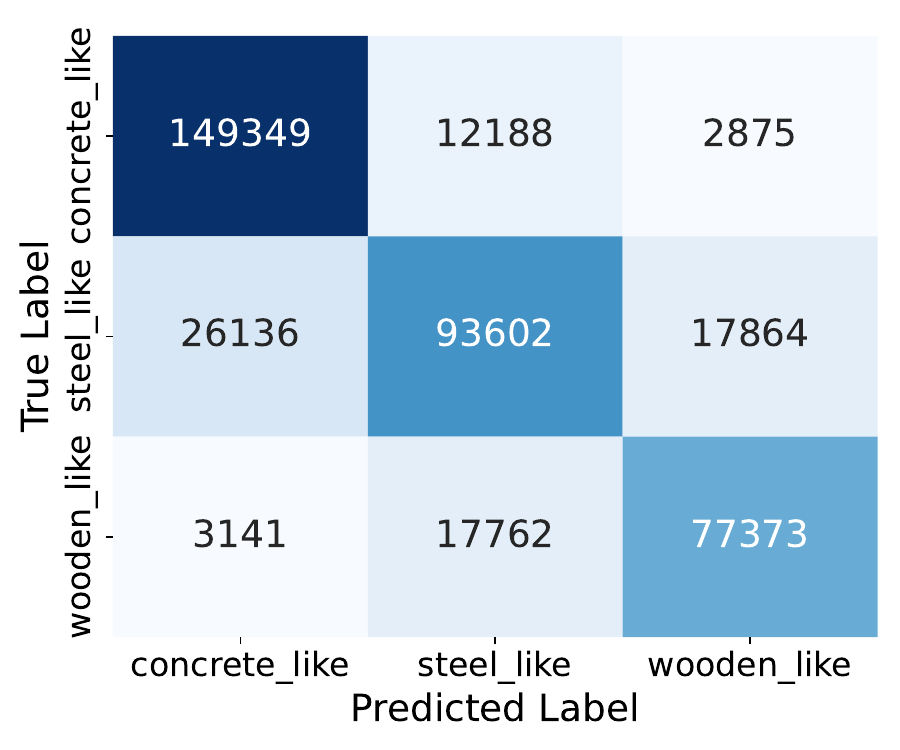}
        \caption{Predicted building structures}
        \label{fig:confusion_matrix_building}
    \end{subfigure}
    \hfill
    \begin{subfigure}[b]{0.27\textwidth}
        \centering
        \includegraphics[width=\linewidth]{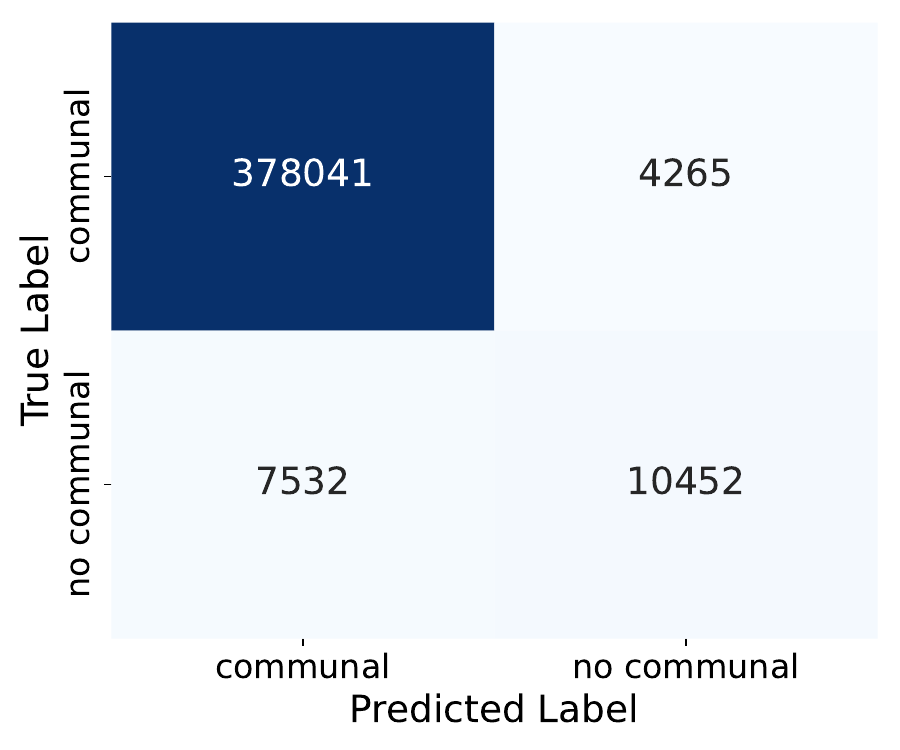}
        \caption{Predicted property types}
        \label{fig:confusion_matrix_property}
    \end{subfigure}
    \hfill
    \begin{subfigure}[b]{0.27\textwidth}
        \centering
        \includegraphics[width=\linewidth]{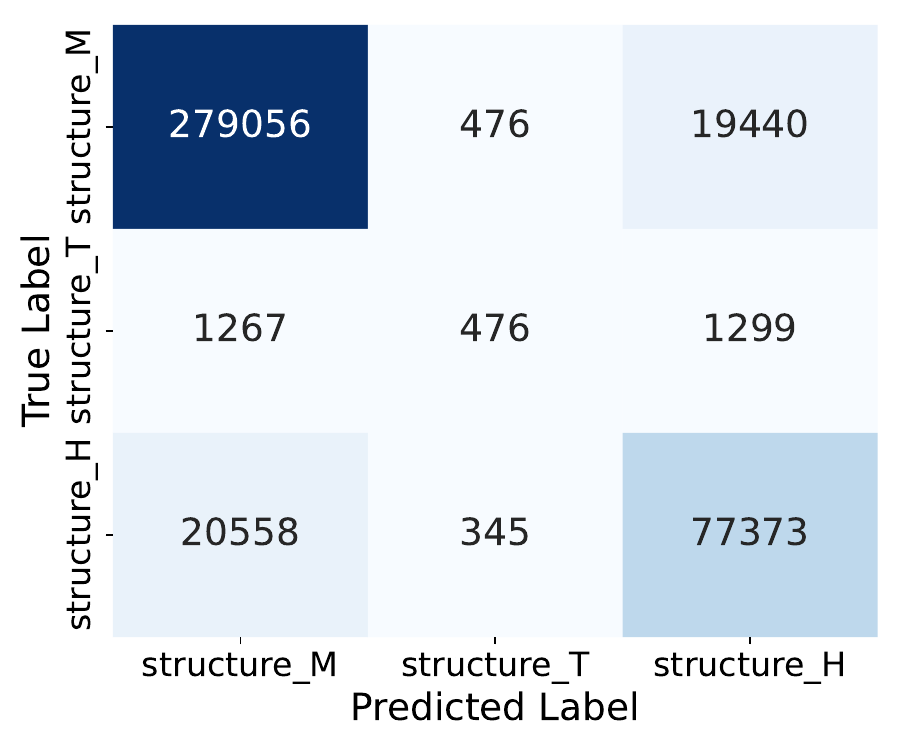}
        \caption{Predicted structural classes}
        \label{fig:confusion_matrix_structure_class}
    \end{subfigure}
    \caption{Intermediate attributes prediction results}
    \label{fig:confusion_matrices}
\end{figure*}

Figures~\ref{fig:confusion_matrix_building} and~\ref{fig:confusion_matrix_property} show the confusion matrices for building-structure and property-type predictions, respectively. 
The building-structure matrix revealed a strong diagonal for concrete-like class, which dominated the training data.
However, a notable differences were observed between steel-like and concrete-like classes. 
This is likely due to visual similarities in facade images and construction features between steel and reinforced concrete buildings, especially in modern multi-unit dwellings.
Moreover, the wooden-like class exhibited high precision and recall relative to its sample size, although underrepresented in the dataset.
This suggests that wooden structures have more distinct visual cues--such as exposed wooden elements or low-rise designs--that are easier to discriminate despite the limited data.

In the property-type confusion matrix, we observe a dominant concentration along the communal axis, reflecting the dataset imbalance toward communal properties. 
Most misclassifications occurred between communal and non-communal types.
When their visual features overlapped, such as between small apartment complexes and detached rental houses. 
These errors are particularly consequential, as the communal versus non-communal distinction directly affects fireproof class derivation for steel-like structures under the rule-based mapping.

Thus, although the overall classification accuracy for both structural type and building category remained high, class imbalance and semantic proximity between categories introduced systematic misclassifications that propagated into incorrect fireproof class labels in the indirect model. 
These findings highlight the importance of intermediate prediction quality, especially in data-sparse or boundary-defining categories.

\begin{table*}[t]
    \centering
    \caption{Performance on intermediate attribute prediction (structural type and building category). 
        Metrics include classification accuracy and macro-averaged F1-score.}
    \label{tab:intermediate_attribute_results}
    \begin{tabular}{ccccc}
        \toprule
        LR & Structure Acc $\uparrow$ & Structure Macro F1 $\uparrow$ & Category Acc $\uparrow$ & Category Macro F1 $\uparrow$ \\
        \midrule
        $10^{-5}$ & 0.9275 & 0.8714 & 0.8321 & 0.7945 \\
        $10^{-6}$ & 0.9182 & 0.8582 & 0.8147 & 0.7721 \\
        \bottomrule
    \end{tabular}
\end{table*}

\subsubsection{Summary of Attribute Prediction Performance (RQ1)}
The results across Tasks 1 to 3 collectively demonstrate that disaster-relevant building attributes can be reliably inferred from facade images using a multi-task learning framework. 
For Task 1, the model achieved a mean absolute error of less than 5 years for construction year estimation, indicating strong predictive capability despite the temporal skew in the dataset. 
In classification tasks (Tasks 2 and 3), the model attained over 92\% accuracy for structural type and over 83\% accuracy for property type, with high macro F1-scores, which reflect robustness against class imbalance. 
These results suggest that the shared visual representation learned by the MTL model captures sufficient architectural cues to support accurate, scalable prediction of building attributes. 
Importantly, all attributes were inferred without relying on handcrafted features or auxiliary metadata, highlighting the feasibility of deploying such models in real-world image-based risk assessment systems.

\subsection{Fireproof Structural Category Prediction (Task 4)}
The ability of the proposed model to predict the fireproof structural category (H, T, and M) was analyzed based on intermediate predictions of building structure and property type. 
This task is particularly relevant in fire and earthquake insurance calculations in Japan.
Table~\ref{tab:fireproof_class_evaluation} summarizes the classification accuracy across two learning rates.
The best-performing configuration (learning rate $10^{-5}$) achieved an overall accuracy of 89.16\%, with a macro-averaged F1 score of 0.6459 and a weighted F1 score of 0.8900.
These results indicate that the proposed model can learn reliable representations for classifying fireproof categories, even when inferred through intermediate attributes.
Per-class F1 scores reveal strong performance for the dominant M class (F1 = 0.9304), and reasonably high performance for the H class (F1 = 0.7880). 
However, classification for the minority T class was challenging, with an F1 score of 0.2194. 
This degradation was primarily attributable to the difficulty of visually distinguishing borderline structural types(e.g. small steel-frame houses vs. large wooden dwellings) and class imbalance--T-type properties accounted for less than 1\% of the training data.

Figure~\ref{fig:confusion_matrix_structure_class} shows the confusion matrix for fireproof category prediction. 
Most misclassifications for T-type structures were confused with the dominant M and H classes, further illustrating the difficulty in separating semi-fireproof buildings based on appearance alone.
Despite these challenges, the attribute-based prediction approach offers several advantages. 
First, the interpretability of intermediate predictions enables transparent error analysis and debugging, which is especially important in regulatory or insurance-related applications. 
Second, rule-based mapping preserves alignment with official fireproof definitions, ensuring consistency with existing insurance frameworks.
The lower-performing configuration ($10^{-6}$) showed uniformly reducing scores across all metrics, suggesting that sufficient gradient updates are essential in this multi-task context. 
In particular, the F1 for class T dropped further to 0.1767, reinforcing the need for targeted strategies (e.g., data augmentation, class reweighting, or cost-sensitive loss functions) to improve minority-class detection.

While the model performed well for the dominant fireproof categories, additional effort is required to improve recognition of semi-fireproof (T) structures—potentially through refining the rule-based labels or rebalancing the training data. 
Nonetheless, the results demonstrate the feasibility of automated, image-based fireproof classification in large-scale residential datasets, providing a foundation for scalable insurance risk profiling.

\begin{table*}[t]
    \centering
    \caption{Performance on fireproof structural category classification (H, T, M). We report accuracy, macro-averaged and weighted F1 scores, and per-class F1 scores to capture both overall and class-specific performance.}
    \label{tab:fireproof_class_evaluation}
    \begin{tabular}{ccccccc}
        \toprule
        LR & Acc $\uparrow$ & Macro Precision $\uparrow$ & Macro Recall $\uparrow$ & Macro F1 $\uparrow$ & Weighted F1 $\uparrow$ & F1 (H/M/T) $\uparrow$ \\
        \midrule
        $10^{-5}$ & 0.8916 & 0.6944 & 0.6257 & 0.6459 & 0.8900 & 0.7880 / 0.9304 / 0.2194 \\
        $10^{-6}$ & 0.8802 & 0.6721 & 0.5998 & 0.6227 & 0.8789 & 0.7692 / 0.9223 / 0.1767 \\
        \bottomrule
    \end{tabular}
\end{table*}

Despite incorrect predictions in intermediate attributes (structure type or property category), the model correctly inferred the structural fireproof class in a significant number of cases.
Approximately 12\% (45,083 instances) of all correctly predicted fireproof labels were achieved despite errors in one or more intermediate attributes, accounting for approximately 11\% of the entire dataset.
This indicates that the model captured meaningful patterns in facade images that can compensate for imperfect intermediate supervision, highlighting the robustness of the hierarchical prediction approach.

These findings collectively address RQ2 and demonstrate that the fireproof structural class can be effectively inferred from intermediate attributes predicted by the multi-task model. 
Although errors in structure type or property category occasionally propagated, the hierarchical approach retained strong predictive performance overall, and even exhibited robustness in cases where intermediate predictions were incorrect.
This suggests that the model leveraged latent visual cues beyond discrete labels, supporting the feasibility of scalable, image-based disaster risk profiling grounded in interpretable intermediate reasoning.

\subsection{Qualitative Examples}
To further interpret the behavior and prediction quality of the model, we present representative facade images whose construction year was predicted with high accuracy (within $\pm 3$ years) across all three fireproof structural categories (H, T, M). 
These examples were drawn from the test set and selected to demonstrate the generalize ability of the model across diverse structural types.

\begin{figure}[t]
    \centering
    \begin{subfigure}[t]{0.14\textwidth}
        \includegraphics[width=\linewidth]{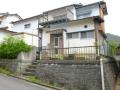}
        \caption{Class H \\ Pred: 1974.37 / True: 1972.0}
        \label{fig:example_H}
    \end{subfigure}
    \hfill
    \begin{subfigure}[t]{0.14\textwidth}
        \includegraphics[width=\linewidth]{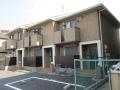}
        \caption{Class T \\ Pred: 2006.20 / True: 2005.0}
        \label{fig:example_T}
    \end{subfigure}
    \hfill
    \begin{subfigure}[t]{0.14\textwidth}
        \includegraphics[width=\linewidth]{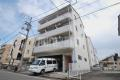}
        \caption{Class M \\ Pred: 1990.67 / True: 1989.0}
        \label{fig:example_M}
    \end{subfigure}
    \caption{Examples of facade images where the construction year was predicted within $\pm 3$ years. Each sample corresponds to a different fireproof structural class.}
    \label{fig:qualitative_HMT_examples}
\end{figure}

As shown in Figure~\ref{fig:qualitative_HMT_examples}, the proposed model successfully identifies subtle visual cues associated with the construction era, even under different structural fireproof classes. 
For instance, in class H (wooden), low-rise designs and traditional architectural elements are present; in class M (reinforced concrete), larger and more modern facades with balconies and uniform cladding are observable. 
The model appears to learn implicit correlations between facade characteristics and construction trends, such as window frame styles or material finishes, which vary by decade.
These examples support the capacity of the model to leverage facade-level patterns for construction-year estimation and demonstrate that accurate predictions can be achieved even in imbalanced or visually diverse categories.

\begin{figure}[t]
    \centering
    \begin{subfigure}[t]{0.14\textwidth}
        \includegraphics[width=\linewidth]{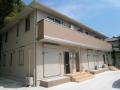}
        \caption{
        \textbf{Pred:} steel\_like / no\_communal $\rightarrow$ T \\
        \textbf{True:} steel\_like / communal $\rightarrow$ M
        }
        \label{fig:misclassified_1}
    \end{subfigure}
    \hfill
    \begin{subfigure}[t]{0.14\textwidth}
        \includegraphics[width=\linewidth]{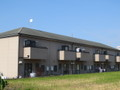}
        \caption{
        \textbf{Pred:} steel\_like / no\_communal $\rightarrow$ T \\
        \textbf{True:} wooden\_like / no\_communal $\rightarrow$ H
        }
        \label{fig:misclassified_2}
    \end{subfigure}
    \hfill
    \begin{subfigure}[t]{0.14\textwidth}
        \includegraphics[width=\linewidth]{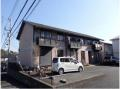}
        \caption{
        \textbf{Pred:} steel\_like / communal $\rightarrow$ M \\
        \textbf{True:} wooden\_like / no\_communal $\rightarrow$ H
        }
        \label{fig:misclassified_3}
    \end{subfigure}
    \caption{
    Examples of misclassified facade images. Each caption shows the predicted and true intermediate attributes (building structure, property type), as well as the resulting structural fireproof classification. Misclassifications often arise due to visually ambiguous features between steel and wooden structures, or between communal and non-communal settings.
    }
    \label{fig:misclassification_examples}
\end{figure}

These examples highlight common sources of error in the indirect model. 
In Figure~\ref{fig:misclassified_1}, the property is a communal steel structure, but was misclassified as non-communal, leading to a downgrade in the fireproof category. 
In Figures~\ref{fig:misclassified_2} and \ref{fig:misclassified_3}, wooden buildings were mistaken for steel, likely owing to cladding or facade renovations that obscure material cues. 
However, our analysis also revealed that a non-negligible subset of samples (approximately 12\% of correctly classified cases) achieved correct fireproof predictions despite inaccuracies in intermediate attributes. 
This highlights both the potential robustness of the indirect model in certain settings and the vulnerability of its hierarchical structure when misclassifications occur, particularly near category boundaries. 
Together, these findings underscore the dual role of intermediate attributes--as a source of interpretability and point of failure--in indirect fireproof classification.

\section{Conclusion}\label{sec:conclusion}
In this study, we proposed an MTL model for predicting disaster-relevant building attributes--namely, construction year, building structure, and property type--from exterior images.
Addressing RQ1, our model demonstrated strong performance across these tasks: for construction-year estimation, the best model achieved an MAE of 4.97 and RMSE of 6.88 years; for structure and property classifications, it reached accuracies of 92.75\% and 83.21\%, respectively. 
Qualitative examples further supported the ability of the model to capture visual cues associated with construction trends and typologies.
To explore RQ2, we applied a rule-based mapping grounded in insurance definitions to derive fireproof structural classes (H, T, M) as an indirect target. 
In this classification task, the model attained an overall accuracy of 89.16\% and a macro F1-score of 0.6459, with particularly high scores for the H (0.7880) and T (0.9304) classes.
Notably, 45,083 instances (approximately 11\% of the dataset, or 12\% of all correctly predicted cases) were correctly classified into fireproof categories despite errors in intermediate attribute predictions, suggesting that the MTL framework supports robust integration of visual signals beyond strict attribute dependency.

The limitations of this study include the use of a lightweight, rule-based estimation system that does not fully capture the complexity of real-world fireproof classification practices. 
In addition, the classification scheme is based on Japanese insurance rating standards, which may limit its applicability in international contexts. 
The model also struggles with underrepresented classes, indicating the need for more balanced or diverse training data.
Building on these limitations, future work will explore end-to-end classification models to reduce error propagation, improve model interpretability through visualization techniques, enhance robustness via advanced data augmentation, and extend applicability by integrating street-level imagery for real-world deployment.

\begin{acks}
In this paper, we used LIFULL HOME'S Dataset provided by LIFULL Co., Ltd. via IDR Dataset Service of National Institute of Informatics.
\end{acks}

\bibliographystyle{ACM-Reference-Format}
\bibliography{sample-base}


\end{document}